%% file: main.tex
\documentclass{article}

\usepackage[preprint]{neurips_2022}

\usepackage[utf8]{inputenc} 
\usepackage[T1]{fontenc}    
\usepackage{hyperref}       
\usepackage{url}            
\usepackage{booktabs}       
\usepackage{amsfonts}       
\usepackage{nicefrac}       
\usepackage{microtype}      
\usepackage{xcolor}         
\usepackage{graphicx}
\usepackage{caption}
\usepackage{card}
\usepackage{cleveref}
\usepackage{multirow}
\usepackage{wrapfig}
\usepackage{floatflt}

\definecolor{green}{HTML}{ccffcc}
\definecolor{yellow}{HTML}{FFFFB3}
\definecolor{blue}{HTML}{B9DEFF}
\definecolor{red}{HTML}{ffcccc}

\title{Paraphrase Detection: Human vs. Machine Content}

\author{%
  Jonas Becker \\
  University of Göttingen\\
  Göttingen, Germany \\
  \texttt{becker@gipplab.org} \\
  \And
  Jan Philip Wahle \\
  University of Göttingen\\
  Göttingen, Germany \\
  \texttt{wahle@uni-goettingen.de} \\
  \AND
  Terry Ruas \\
  University of Göttingen\\
  Göttingen, Germany \\
  \texttt{ruas@uni-goettingen.de} \\
  \And
  Bela Gipp \\
  University of Göttingen\\
  Göttingen, Germany \\
  \texttt{gipp@uni-goettingen.de} \\
}

\begin{document}

\maketitle

\begin{abstract}
  The growing prominence of large language models, such as GPT-4 and ChatGPT, has led to increased concerns over academic integrity due to the potential for machine-generated content and paraphrasing. Although studies have explored the detection of human- and machine-paraphrased content, the comparison between these types of content remains underexplored. In this paper, we conduct a comprehensive analysis of various datasets commonly employed for paraphrase detection tasks and evaluate an array of detection methods. Our findings highlight the strengths and limitations of different detection methods in terms of performance on individual datasets, revealing a lack of suitable machine-generated datasets that can be aligned with human expectations. Our main finding is that human-authored paraphrases exceed machine-generated ones in terms of difficulty, diversity, and similarity implying that automatically generated texts are not yet on par with human-level performance. Transformers emerged as the most effective method across datasets with TF-IDF excelling on semantically diverse corpora. Additionally, we identify four datasets as the most diverse and challenging for paraphrase detection.
\end{abstract}

\section{Introduction}

Paraphrase detection plays a crucial role in maintaining the integrity of scholarly works and written content in general. The ability to identify semantically similar texts, despite significant differences in word choice and structure, is essential for various applications, including plagiarism detection \citep{FoltynekMG19} and evaluating the discrepancy between machine-generated and human-generated paraphrases. The growing popularity of large language models (LLMs), such as GPT-4 \citep{gpt4technicalreport} and ChatGPT, has enabled the automatic generation of high-quality paraphrases \citep{WahleRFM22, DouFKS22}, further emphasizing the need for robust detection methods.

While research in natural language processing (NLP) has extensively explored paraphrasing by both humans \citep{kovatchev-etal-2019-qualitative,articleSemSim} and machines \citep{ChenTXH20,MengAHS21a,MaddelaAX21,YangLLY22}, there is a notable lack of studies comparing human-generated paraphrases to those produced by machines \citep{WahleRKG22a}. Gaining a deeper understanding of their similarities and differences is crucial for improving paraphrase detection techniques and mitigating potential threats to academic integrity.

Alignment research, which seeks to identify semantic correspondences between different textual representations, holds promise for enhancing our understanding of machine-generated and human-generated paraphrases \citep{burdick-etal-2022-using}. Investigating the patterns that machines can generate effectively compared to humans can inform data augmentation strategies \citep{kumar-etal-2019-submodular, cao-etal-2020-unsupervised-dual}, which can be particularly valuable given the limited availability of positive training examples for plagiarism detection systems \citep{FoltynekMG19}.

In this paper, we elucidate the similarities and differences between machine and human paraphrases. We assess the performance of seven automated paraphrase detection methods, ranging from classical approaches (e.g., N-Gram) to modern Transformers (e.g., BERT), on aligned text pairs. Additionally, we provide a comprehensive overview of 12 paraphrase datasets (both machine- and human-generated) and highlight the critical attributes to consider when selecting a dataset for research purposes. Alongside our experiments, we publish the source code for reproducability\footnote{\url{https://github.com/jonas-becker/pd-human-vs-machine-content}}.

Our technical report has the following main findings:
\begin{itemize}
    \item Machine paraphrases are semantically closer than human paraphrases.
    \item Most detection methods can identify machine-generated data easier.
    \item ETPC, APT, TURL and QQP are the most diverse and semantically challenging datasets.
    \item Transformers are generally outperforming other detection methods but TF-IDF cosine distance still proves to be a very strong baseline.
\end{itemize}

The implications of our study on paraphrase detection extend across various domains and applications. As the availability of positive training examples for plagiarism detection systems is typically limited due to the scarcity of known plagiarism cases \citep{FoltynekMG19}, our research holds significant value. By identifying and understanding the paraphrase patterns generated effectively by machines in comparison to humans, we can facilitate data augmentation strategies \citep{kumar-etal-2019-submodular, cao-etal-2020-unsupervised-dual}, ultimately enhancing the performance of detection systems. Furthermore, recognizing instances where machine-generated paraphrases closely resemble human-written originals enables the learning of semantic representations in large LLMs \citep{burdick-etal-2022-using}, which can lead to improvements in various NLP tasks, including text summarization \citep{kirstein-etal-2022-analyzing}, and sentiment analysis \citep{zhang2021aspect}. One of our main findings that shows how machine paraphrases are less challenging to detect than their human-generated counterparts gives hope in the massive landscape of machine-generated texts by LLMs like ChatGPT and GPT-4.

\section{Related Work}
Paraphrases are texts that convey approximately the same meaning with different wording and have been used for many NLP tasks, such as plagiarism detection \citep{FoltynekRSM20a,WahleRMG21,WahleRKG22a}, machine translation \citep{thompson-post-2020-automatic}, or word sense disambiguation \citep{WahleRMG21a}.

Paraphrasing is one of the core detection challenges in plagiarism identification. 
\citet{Chow} provided a taxonomy of plagiarism detection methods and their classification according to detection ability for a specific type of plagiarism (i.e., with increasing complexity: copy, near copy, restructuring, paraphrasing, summarizing, translating). 
These range from the finding of context similarities via n-grams \citep{Kond} over the comparison of semantic classes of words \citep{Resn} to syntactical comparisons \citep{uzuner-etal-2005-using}. 
They found semantic-, fuzzy-, and n-gram-based to be the best methods for detecting strong paraphrases. Therefore, this study focuses on these three plagiarism detection methods.

\textbf{Human Paraphrases.} \citet{mehdizadeh-seraj-etal-2015-improving} used multilingual paraphrasing to provide translations for out-of-vocabulary phrases. 
\citet{dong-etal-2017-learning} presents a general framework to learn felicitous paraphrases for question answering in Freebase.
Recently, Transformers \citep{attentionisallyouneed} dominate the field of human paraphrase identification \citep{devlin-etal-2019-bert, xlnet, raffel2020exploring}. Transformer networks in the NLP field generally contain an encoder and a decoder. The encoder is used to create the embedding (high-dimensional vector representation of a text). The decoder aims at reconstructing text by taking the embedding as a semantic representation. An optimal Transformer provides a minimal reconstruction loss between the original text and the decoded embedding \citep{attentionisallyouneed}.
Downstream benchmarks commonly include paraphrase detection as one of the model's desired abilities for solving natural language understanding (e.g., MRPC in GLUE \citep{wang2018glue})

\textbf{Machine Paraphrases}. \citet{Foltynek2020} evaluated the effectiveness of several word embedding and machine learning techniques in identifying machine paraphrases of online paraphrasing tools. Combining word2vec \citep{word2vec} embedding models and an SVM classifier could outperform human experts on examples from Wikipedia, arXiv, and student theses. 
\citet{WahleRFM22} extended this study by evaluating neural language models, such as BERT and Longformer, outperforming previous methods with superhuman performance.
When composing paraphrases with large language models such as T5 or GPT-3, humans were largely unable to detect generated content anymore \citep{WahleRKG22a}.

Despite the progress in paraphrase detection methods, there remains a gap in the literature regarding the comprehensive evaluation and comparison of state-of-the-art models in diverse scenarios. While many studies have contributed to the development and application of advanced Transformer models like BERT and T5, a systematic comparison highlighting their strengths, weaknesses, and best practices for various paraphrase detection tasks is still lacking. Furthermore, the rapid evolution of NLP models necessitates a continuous reassessment of their performance in relation to paraphrase detection. Our study aims to fill this gap by providing an in-depth analysis of detection models, exploring their performance in different detection scenarios, and addressing the questions that previous studies have not yet fully answered.

\section{Methodology}

\subsection{Datasets}

We selected seven datasets composed of human-generated paraphrases and five datasets with machine-generated paraphrases. 
We selected datasets from tasks listed at ``Papers With Code'' and added auxiliary tasks from related works.

\subsubsection{Human-Generated}
\indent\indent\textbf{ETPC.} The \textit{Extended Paraphrase Typology Corpus} is an extension of the MRPC corpus \citep{dolan-brockett-2005-automatically} composed of 5k human-written newswire articles. ETPC adds annotations for 27 paraphrase types according to their typology \citep{etpc}. 

\textbf{QQP.} The \textit{Quora Question Pairs} corpus contains 400k similar questions from Quora with binary type annotations of whether a question is a paraphrase of another \citep{QQP}.

\textbf{TURL.} The \textit{Twitter News URL Corpus} contains 2.8m human-generated sentence pairs from Twitter news with binary type annotations of six human raters \citep{lan-etal-2017-TURL}. As TURL was the only dataset considered with multi-rater annotations, we labeled a paraphrase positive using a majority vote and disregarded pairs without majorities.

\textbf{SaR.} The \textit{IBM Split and Rephrase} dataset contains 2k complex sentences from Wikipedia and legal documents that have been simplified by human annotators, hence splitting the sentence into parts and rephrasing its content \citep{zhang-etal-2020-small}.

\textbf{MSCOCO.} The \textit{Common Objects in Context} dataset consists of 328k images that each have been annotated by multiple human participants \citep{mscoco}. We sample one random annotation pair of an image as paraphrased pair as it is common practice amongst other paraphrase studies \citep{fu2020paraphrase}.

\textbf{ParaSCI.} The \textit{ParaSCI} dataset contains 350k text pairs from arXiv and ACL Anthology with long divergent sentences (approximately 19 words) \citep{ParaSCI}.

\subsubsection{Machine-Generated}

\indent\indent\textbf{MPC.} The \textit{Machine Paraphrase Corpus} dataset contains over 200k machine-paraphrased paragraphs from Wikipedia, arXiv, and student theses using BERT and a masking probability of 30\% \citep{WahleRFM22}.

\textbf{SAv2.} The \textit{Sentence Simplification} dataset contains 60k machine-generated text pairs from English Wikipedia that are paraphrased by leaving out information and truncating sentences \citep{kauchak-2013-improving}.

\textbf{ParaNMT-50M.} The \textit{Paraphrase Neural Machine Translation} dataset contains 50m machine-generated sentence pairs that have been obtained by back-translating the Czeng1.6 Corpus \citep{Bojar2016CzEng1E} to English \citep{wieting-gimpel-2018-paranmt}.

\textbf{PAWS-Wiki.} The \textit{Paraphrase Adversaries from Word Scrambling} dataset contains 65k machine-generated text pairs in its Wikipedia split that have been acquired by performing word-swapping and back-translation \citep{PAWSDataset}.

\subsubsection{Human- \& Machine-Generated}
\indent\indent\textbf{AP.} The \textit{Adversarial Paraphrasing} dataset contains two subcorpora. The \textit{APH} split includes 4975 human-generated pairs from the MSRP corpus and an updated google-translated version of the ParaNMT dataset (called PPNMT \citep{nighojkar-licato-2021-improving}). The \textit{APT} split comprises data that has been generated using T5 to machine-paraphrase texts from the MSRP dataset leading to a total of 138k sentence pairs \citep{nighojkar-licato-2021-improving}. Both subsets are evaluated individually in this study.

\subsection{Detection Methods}
\label{chap:detectionmethods}

Our proposed pipeline incorporates a diverse range of metrics, encompassing both classical approaches and modern Transformer techniques, as inputs to a Support Vector Machine (SVM) for text-matching. We will hereafter refer to this process of using a method as input to an SVM as the detection method. To facilitate a comprehensive comparison between datasets and the advantages and disadvantages of various detection methods, we have selected a combination of baseline methods, and advanced alternatives, as suggested by \citet{Chow}.
For further information on the specific methods employed, refer to \Cref{tab:table-detection-methods}.

\input{tables/method_comparison}

\section{Experiments}

\subsection{Preprocessing}
\Cref{tab:parsed-datasets} outlines how we preprocess the datasets before applying detection. 
The top datasets contain human-generated paraphrases and the bottom ones machine-generated paraphrases. 
We filtered out pairs with empty texts or where both texts were equal (``Filtered''). 
``Duplicates'' are text pairs where either the first text or the second text was found in another pair's first text or second text. 
We counted filtered and duplicate pairs up until 10k pairs from each dataset were retrieved as the maximum amount per dataset.
The datasets TURL and SAv2 required the most filtering of unusable pairs.
On the contrary, ETPC, SaR, and APT showed to contain the most diverse, and therefore usable pairs. 

Some datasets did only contain positive examples (true paraphrases). To balance out their data for training and testing, we supplemented these datasets with i.i.d. randomly sampled negative pairs. 
Together with the supplements, we balanced pairs to contain 50\% paraphrased and 50\% original text pairs.

To prevent our final model from being biased towards specific datasets, we restrict the training corpus for each dataset to a maximum of 4k pairs. The sole exception to this limitation is the SaR dataset, which has an inherently small corpus.

\input{tables/parsed_datasets.tex}

\subsection{Detection Method Setup}
We use each detection method for training an SVM as they provide a unified, yet inexpensive framework to compare different features like n-grams to Transformers \citep{brockett2005support}. 
The methods TF-IDF, N-Gram, and Fuzzy return a float similarity score that described the similarity (higher means more similar). 
GloVe and Fasttext give a vectorized variant of both texts within each pair. 
For BERT and T5 we use the embedded texts as our detection metric.
We provide more details about the detection methods in \Cref{chap:detectionmethods}.

\subsection{SVM Training}
For a balanced training process, we train an SVM across all datasets once per detection method.
To find suitable parameters for the SVM tailored to each method, we apply a small randomized grid search of 25 iterations with 2-fold cross-validation \citep{RuasFGF20,WahleRMG21a,FoltynekRSM20a} (shown in \Cref{tab:table-gs-params}). 
We perform a grid search seven times (once per method) and choose the trained SVMs with the highest test scores across datasets for further analysis.

\input{tables/gs_params}

\subsection{Similarity Analysis}
A comparison is conducted for all 12 datasets with regard to their semantic similarity and overall thematic structure. BERT is utilized to generate embeddings for all pairs. Following this, the embedding space of BERT's \texttt{[CLS]}-embedding is reduced to two dimensions using t-distributed stochastic neighbor embedding (t-SNE), preserving the local structure between the datasets' text pairs. In addition to our visual analysis, the mean cosine distance of the BERT embeddings is computed to illustrate the similarity or dissimilarity of pairs within each dataset, further supporting the qualitative aspect of our investigation.

\section{Evaluation}

\begin{wrapfigure}[15]{r}{9cm}
    \vspace*{-.45cm}
    \begin{minipage}{.5\textwidth}
    \centering
    \includegraphics[width=7cm]{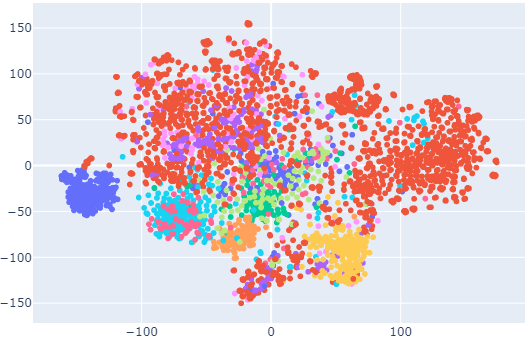}
    \end{minipage}%
    \begin{minipage}{.5\textwidth}
    \hspace*{0.2cm}
      \includegraphics[width=.25\linewidth]{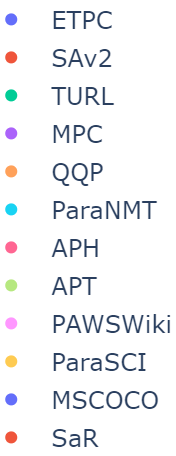}
      \label{fig:test2}
    \end{minipage}
    \caption{A two-dimensional t-SNE representation of the datasets' text embeddings.}
    \label{fig:embedded}
\end{wrapfigure}

We first evaluate the t-SNE visualized BERT embeddings of all texts (\Cref{fig:embedded}). APH and APT were sampled partially from the same data source and paraphrased by humans and T5, respectively.
Their semantic representations are close but do not overlap.
MPC has multiple distinct clusters of many sources (Wikipedia, arXiv, student theses) and topics. 
SAv2's texts also result in distinct clusters even though they all come from the same source (Wikipedia).

The observed clusters indicate that the selected datasets have thematic diversity. SaR returned a very dense and distinct secondary cluster that mainly included intro sentences from legal documents. 
ParaNMTs scatter was broader, with one primary and another secondary cluster. 
ETPC, APT, TURL, and QQP showed thematically as the most balanced datasets in our selection with a broad and relatively even scatter.

When looking at the mean cosine distances for paraphrased text pairs only, it becomes clear that machines generally create less diverse paraphrases in our dataset selection (e.g., MPC often just changes a single word in a pair). 
This indicates that machine paraphrases are easier to detect using semantically based approaches compared to human paraphrases. However, we need to consider that the MPC dataset specifically used BERT to generate its paraphrases. Thus, the semantic representations (BERT embeddings) of its pairs are expected to be close. Yet, other machine-generated datasets were formed using other methods and still show a higher cosine similarity compared to human-generated datasets.

Our findings also reveal that ETPC (0.93 vs 0.90 cosine similarity) and PAWS-Wiki (0.98 vs 0.98 cosine similarity) exhibit the smallest difference between paraphrased and original pairs. This factor significantly impacts the reliability of a model trained and tested on these datasets individually when encountering unseen data. For instance, models trained on datasets like TURL, which exhibit a high difference, will achieve promising results quickly; however, they may demonstrate significantly lower recall in detection tasks when applied to datasets like PAWS-Wiki or ETPC, which have a low difference.

\subsection{Paraphrase Detection}

We evaluate each detection method by its F1-score of the SVMs test prediction. \Cref{tab:table-methods-per-dataset-eval} shows the results. Overall, BERT is the strongest method in detecting aligned paraphrased text pairs.
BERT achieves the best F1-scores across most human-paraphrased examples as well as for machine-paraphrased data. Moreover, BERT and T5 clearly show better performance on datasets that displayed distinct semantic clustering (\Cref{fig:embedded}) implicating that detecting paraphrases on more diverse data is increasingly challenging. 
As a result, more classical approaches (e.g., using the TF-IDF cosine distance as an input to an SVM) perform better than expected on topically and lexically diverse corpora.

Generally, all methods were able to detect machine paraphrases. 
The lowest average F1-score across methods was achieved on the human-generated TURL dataset (0.48).

For human datasets, SaR was the easiest to detect (F1: 0.84). 
The construction method of SaR (human annotators splitting the sentence and applying rephrasing to sentence pieces) leads to paraphrases being semantically and textually very close.
This finding is supported by BERT's high cosine similarity of 0.97 (\Cref{tab:cosine_sim}).
Even though ETPCs' paraphrases displayed to be semantically challenging to detect, the lion's share of methods still performed well on ETPC.

\begin{wraptable}[16]{r}{5.5cm}
    \vspace*{-.45cm}
    \caption{Mean cosine similarities of text pairs embeddings (paraphrases and originals separately).}
    \centering
    \input{tables/mean_cos_distances_embedded.tex}
    \label{tab:cosine_sim}
\end{wraptable}

Across machine datasets, we find the MPC dataset to contain a lot of text pairs that are easier to detect for the majority of methods (TF-IDF being the exception).
We achieved an F1-score of 0.86.
In practice, the MPC dataset contains weak paraphrases (e.g., single-word changes or punctuation changes).
This also correlates with the high mean cosine similarity (0.99) of MPCs' BERT embeddings (\Cref{tab:cosine_sim}).

N-Gram performed worst across most datasets. APT is a noteworthy exception to this as it is a dataset acquired by utilizing T5 as a perturbation model. 
For the MSCOCO dataset, N-Gram fails (F1: 0.09) at catching semantic similarities for independent image descriptions with varying vocabulary and levels of detail.

\input{tables/methods_per_dataset_eval.tex}

\subsubsection{Correlations}
\begin{figure}[t]
    \centering
    \includegraphics[width=.45\linewidth]{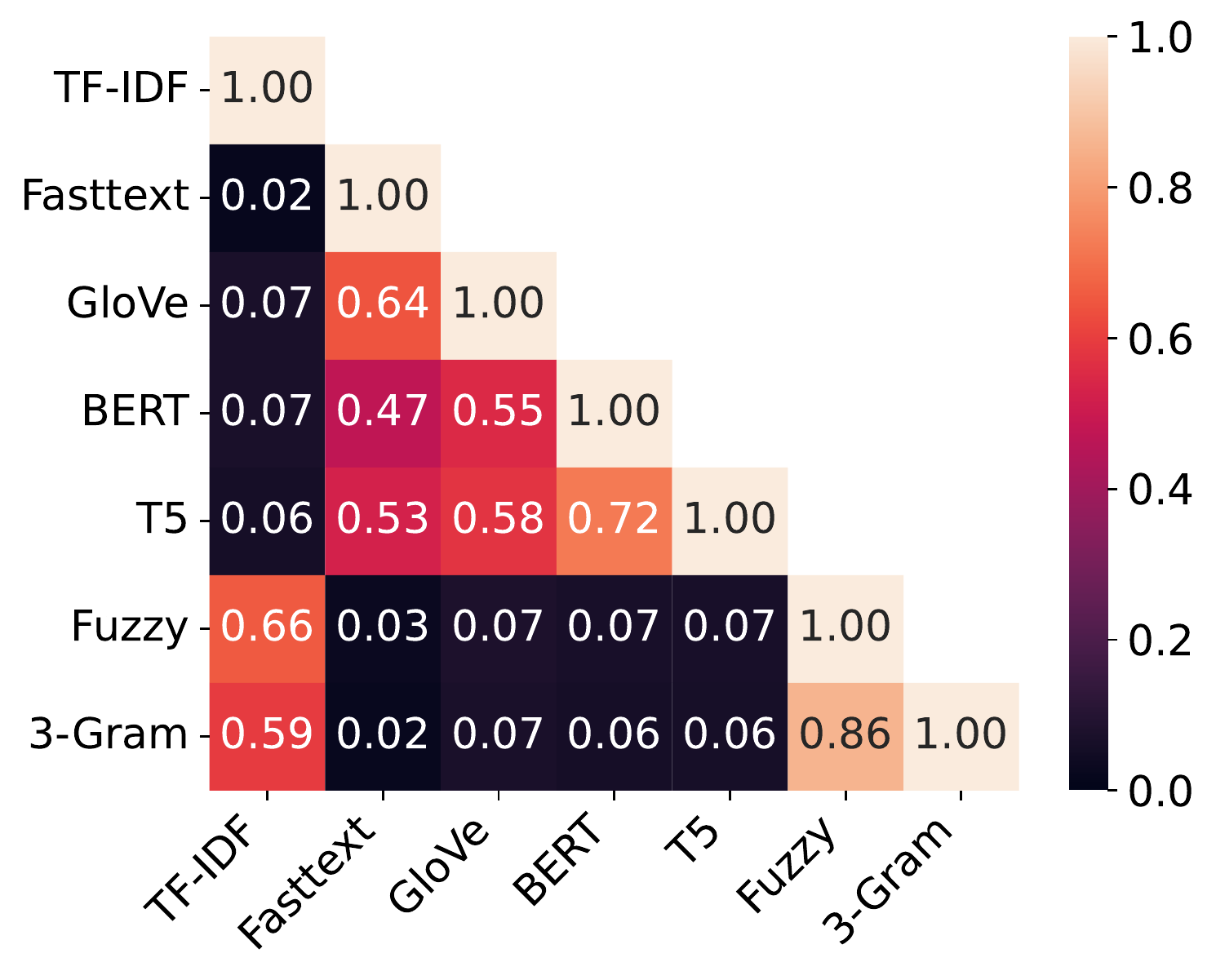}
    \hspace{0.5cm}
    \includegraphics[width=.45\linewidth]{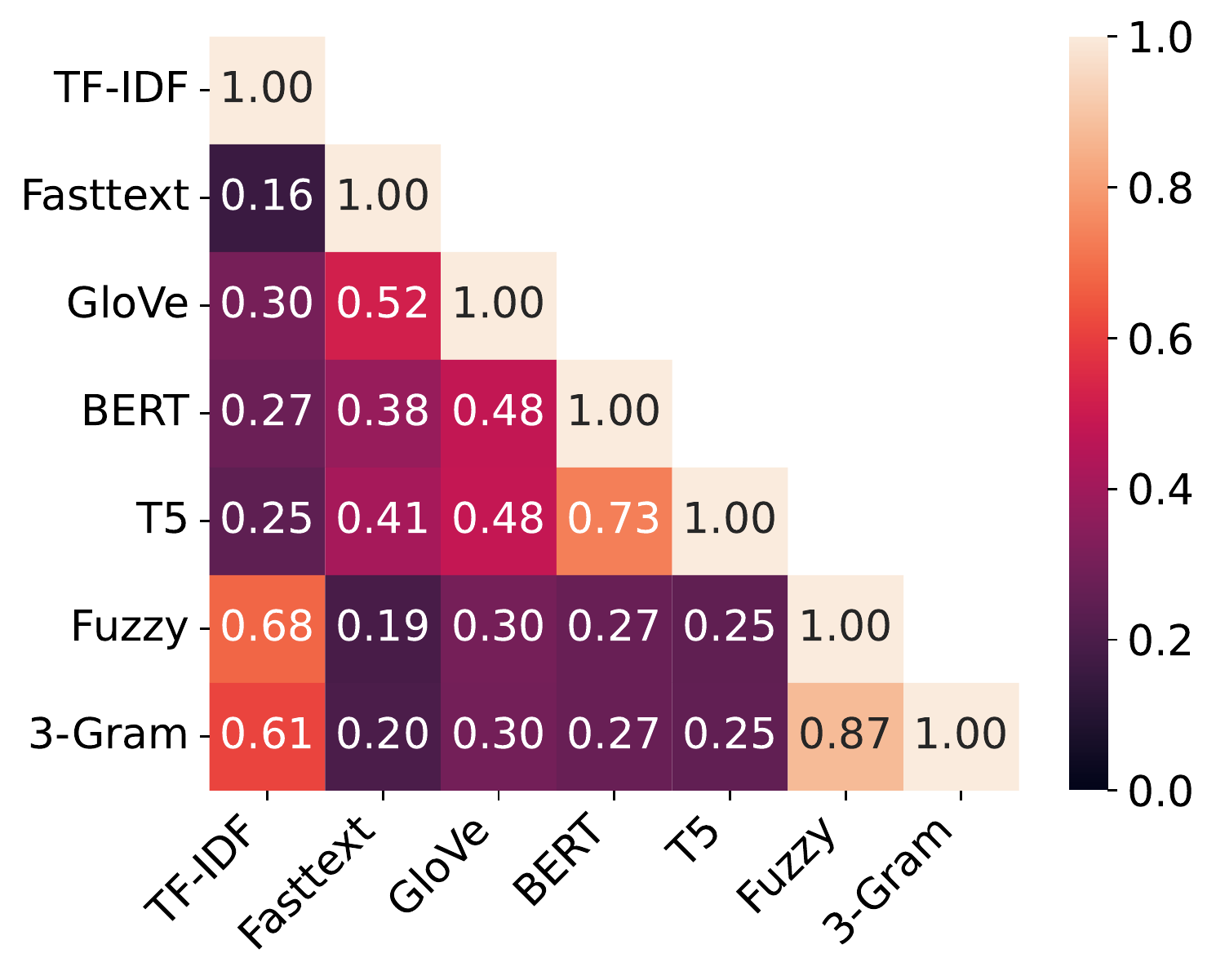}
    \caption{Correlations of detection methods for human paraphrases (left) and machine paraphrases (right). \label{fig:correlation_heatmap}}
\end{figure}

\Cref{fig:correlation_heatmap} displays the correlations of all detection methods as a heatmap with human paraphrases on the left and machine paraphrases on the right. 

Correlations are significantly stronger between many methods of machine-generated data than for human-generated text.
This means that there exists a much higher consensus between these methods which is in line with our findings of the F1-scores. Scores are noticeably improved on machine datasets as seen in \Cref{tab:table-methods-per-dataset-eval} with the average F1-score across human datasets and machine datasets at 0.65 and 0.71 respectively.

We found that BERT and T5 demonstrate the strongest correlations when applied to machine-generated data. Both models exhibit high performance on human-generated data, as evidenced in \Cref{tab:table-methods-per-dataset-eval}. However, when analyzing machine-paraphrased data within our aligned scenario, the improvement in performance for BERT and T5 is marginal.

While TF-IDF, N-Gram, and Fuzzy matching did not yield higher correlations between machine and human data, they still demonstrated relatively high correlations overall. This observation suggests that these methods may share similarities in detection tasks due to their word-matching nature. The observed low correlations between TF-IDF, N-Gram, and Fuzzy matching and Fasttext, GloVe, BERT, and T5 respectively provide some insight into their differences across tasks and generation methods, but further research is needed to confirm these results.

\begin{wraptable}[15]{r}{3cm}
    \vspace*{-.45cm}
    \caption{Gini Coefficients of SVMs predicted probabilities across all datasets.}
    \centering
    \input{tables/gini_coefficients}
    \label{tab:table-gini-coeffs}
\end{wraptable}

The correlation between TF-IDF and Fasttext is found to be lowest for both human and machine-generated content. This can be attributed to the fact that Fasttext represents words as probability distributions, which take into account the context in which they appear, while TF-IDF relies on word frequency and calculates vector cosine distances. Additionally, Fasttext also captures N-Grams in its probability distributions, which allows it to capture more nuanced relationships between words. This feature is difficult for TF-IDF to extract, as it does not consider the ordering of words in a sentence. Therefore, the lower correlation between TF-IDF and Fasttext can be attributed to the more sophisticated contextual representation that Fasttext provides, which is not captured by TF-IDF.

\begin{figure}[!ht]
    \centering
    \includegraphics[width=.45\linewidth]{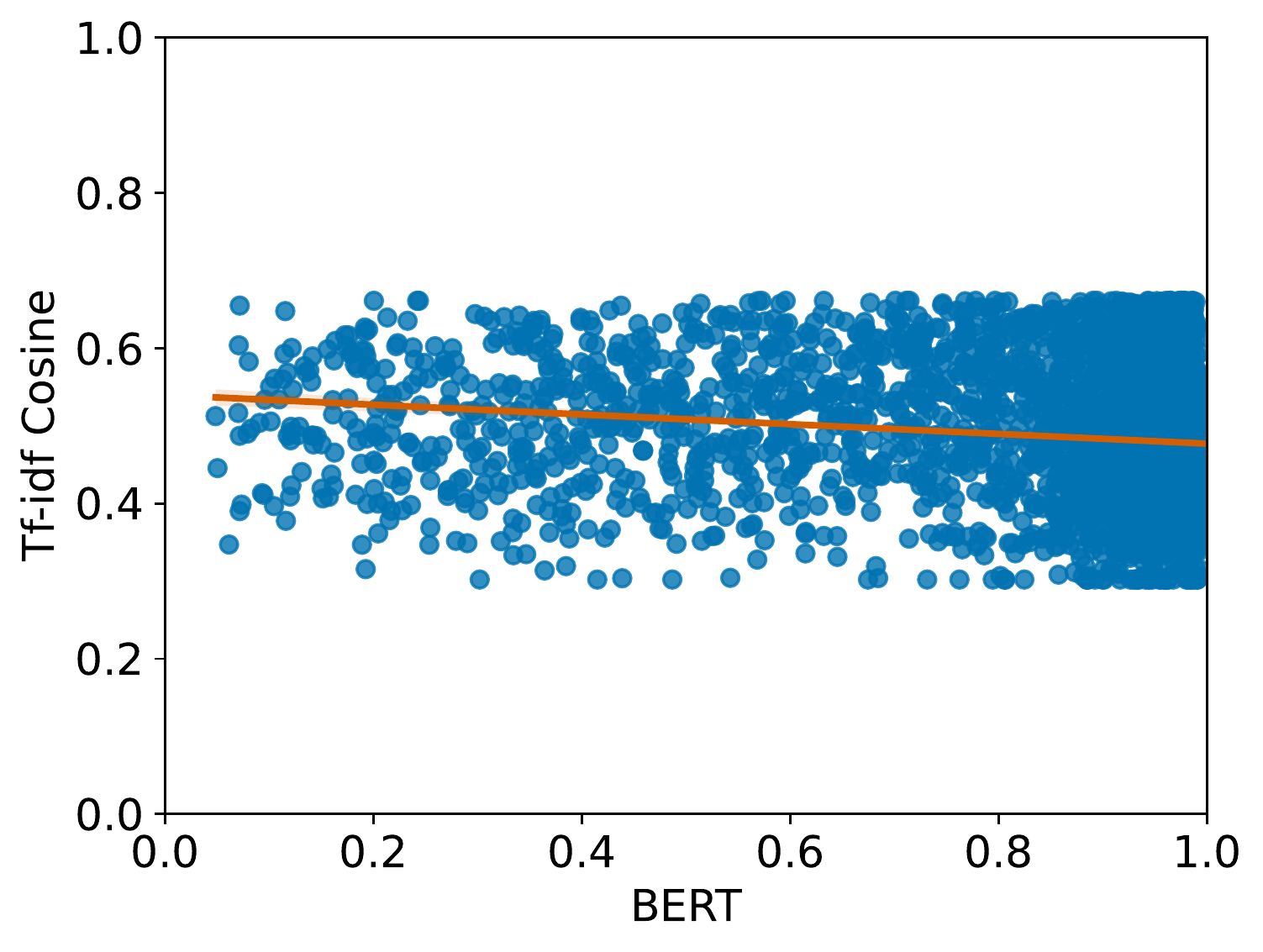}
    \hspace{0.5cm}
    \includegraphics[width=.45\linewidth]{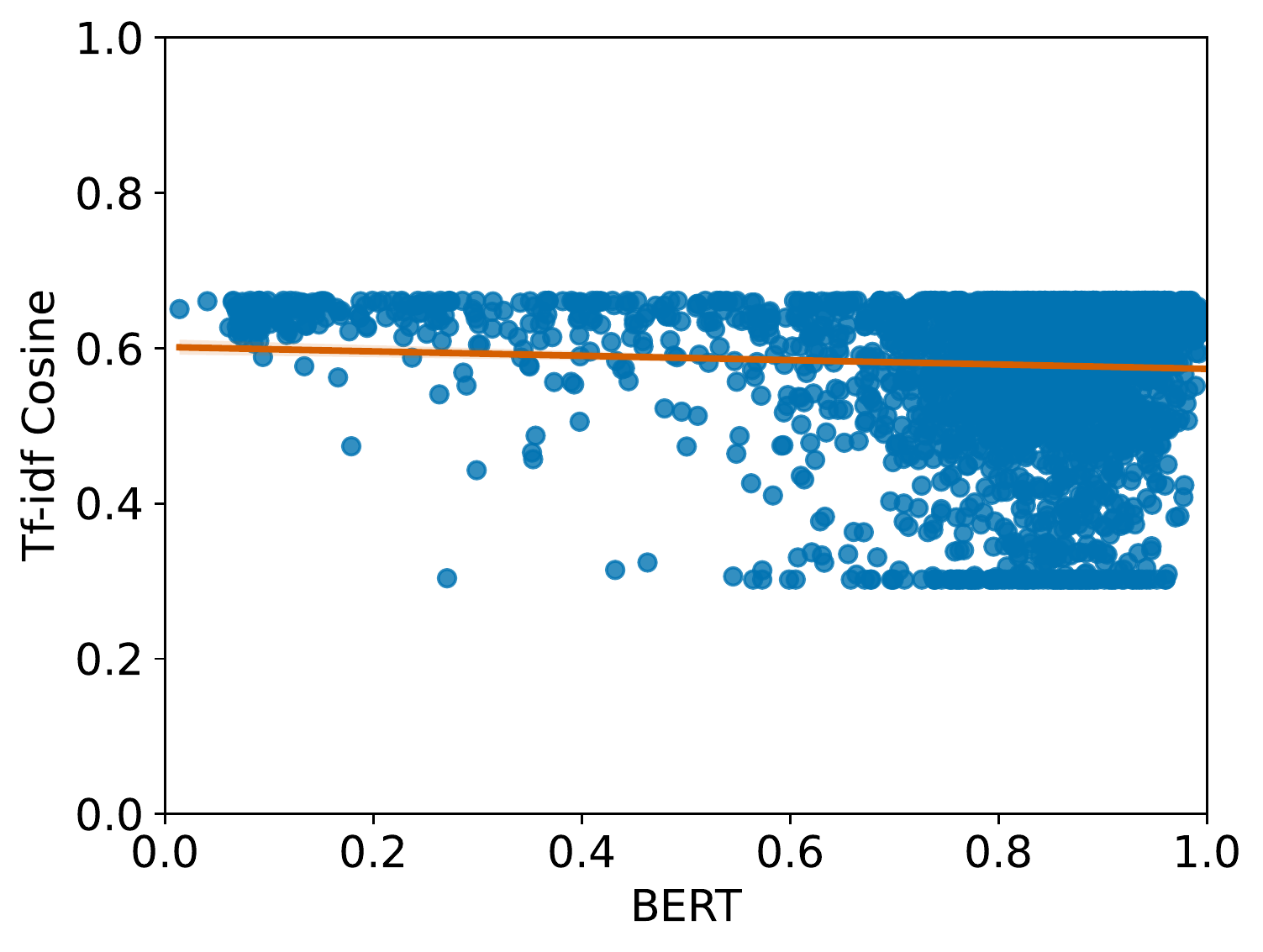}
    \includegraphics[width=.45\linewidth]{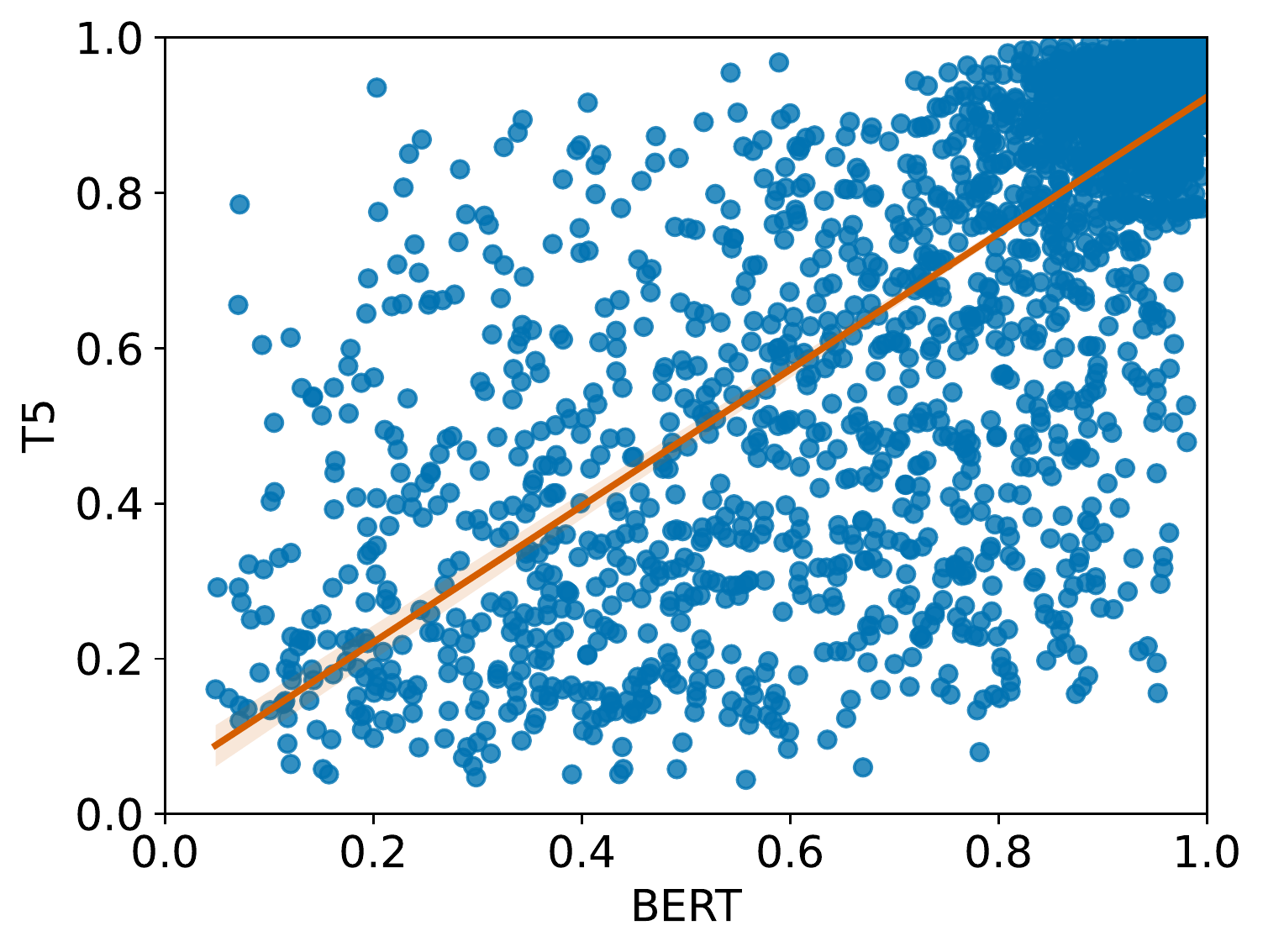}
    \hspace{0.5cm}
    \includegraphics[width=.45\linewidth]{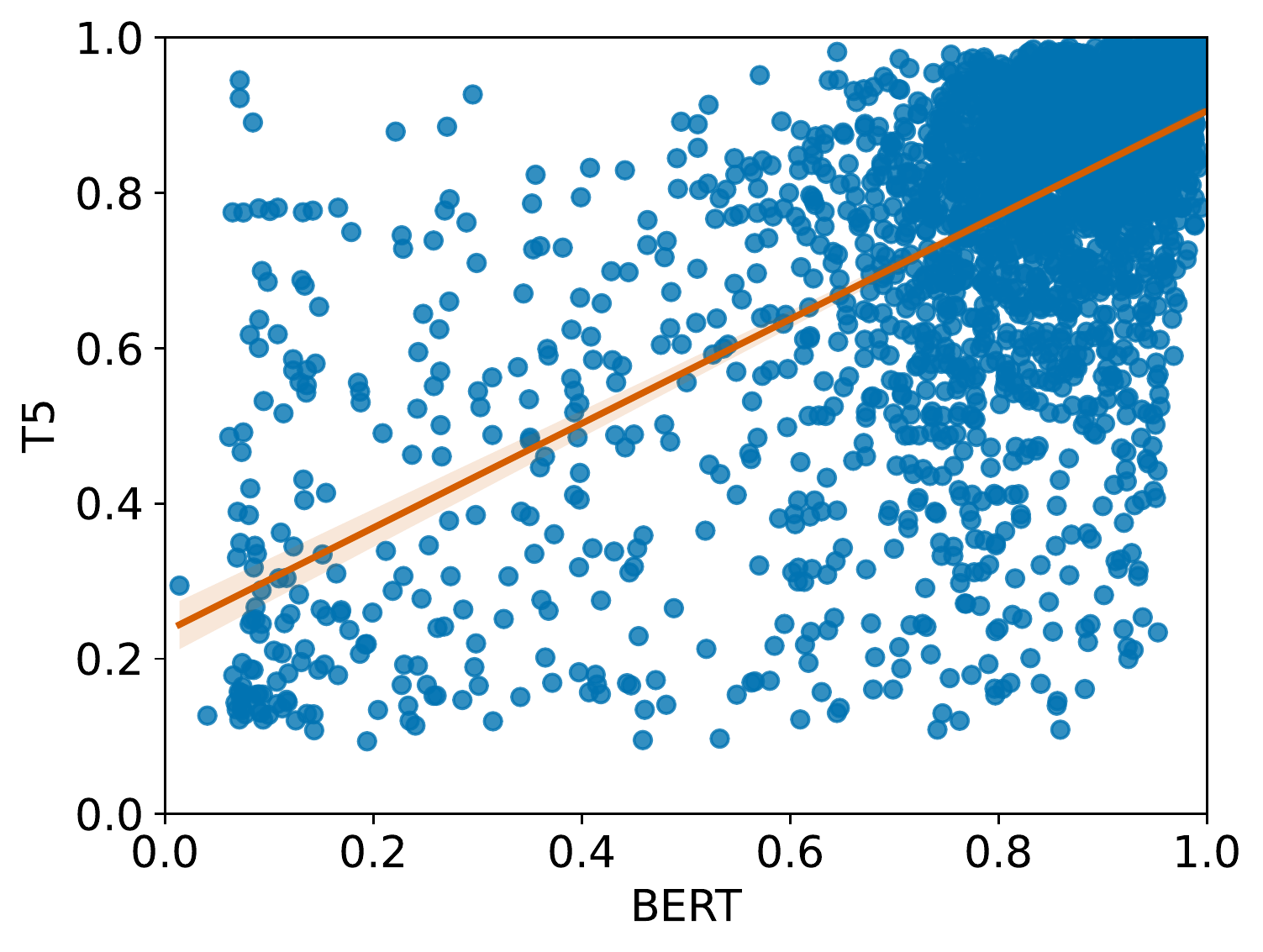}
    \caption{One-on-one correlation graphs on the SVMs predicted probabilities of human paraphrases (left) and machine paraphrases (right). \label{fig:correlation_one_on_one}}
\end{figure}

\Cref{fig:correlation_one_on_one} selectively displays one-on-one correlations between two detection methods with human paraphrases on the left and machine paraphrases on the right. 
Each point is the model's predicted probability to be a paraphrase.

We found that the predicted probabilities of BERT and T5 are present within a much larger range than other methods (nearly using the whole interval of probabilities $[0,1]$). TF-IDF, N-Gram, and also Fuzzy matching are limited to $[0.3,0.8]$, making them more concentrated. We validate the diversity of predictions by calculating the Gini Coefficient \citep{GiniCoefficient} for each detection method's predictions (\Cref{tab:table-gini-coeffs}).
The Gini Coefficients confirm that approaches like TF-IDF (0.10) yield much less diversity compared to Transformers like BERT (0.36) supporting the observed concentrated probability distributions.

\section{Epilogue}
\subsection{Conclusion}

Paraphrasing is an essential aspect of natural language understanding, and studying the differences between human-generated and machine-generated paraphrases is becoming increasingly important. This is particularly relevant given recent innovations in models like ChatGPT and GPT-4, which are capable of generating highly realistic paraphrases.

In our analysis, we examined various approaches to paraphrase detection and assessed a range of datasets commonly used in the field. We observed that not all datasets offer equal success in providing high-quality paraphrase pairs, with machine-paraphrased datasets often lacking diverse pairs. Some datasets, such as SAv2 and MPC, displayed imbalances in semantic aspects, highlighting the need for diverse training sets when developing paraphrase detection methods.

Among the datasets we evaluated, ETPC, APT, TURL, and QQP emerged as the most thematically balanced. Furthermore, the mean cosine distance of BERT embeddings indicated that ETPC and PAWS-Wiki presented the highest prediction task difficulty.

Our experiments demonstrated that modern Transformer approaches outperform traditional methods. However, employing TF-IDF cosine distance on lexically diverse corpora still yielded strong results. Transformers showed their strength in detecting paraphrases on thematically imbalanced datasets, with BERT and T5 providing higher certainty in detection due to a significantly larger range of predicted probabilities during testing.

In conclusion, current machine-paraphrasing datasets are often less challenging for advanced detection methods than human-generated datasets in the aligned case. This underscores the need for continued development of high-quality machine-generated training data that includes a substantial number of strong paraphrases and exhibits both lexical and semantic balance.

\subsection{Limitations}
This study has the following potential limitations. Due to limited computing resources, we capped the total size of nine out of 12 considered datasets to a maximum of 10k examples. The random sampling of the drawn examples might not represent the entire dataset which can bias parts of our analysis. However, using our open-source implementation, one can reproduce the experiments with the full datasets if computational resources are available.

The array of datasets we used for our experiments covered corpora that did not include negative pairs (i. e. only paraphrased pairs). Due to the lack of diverse datasets particularly for machine-generated texts, we supplemented these corpora with random negative pairs from other datasets. Nonetheless, the detection performance of some datasets may be influenced by, e.g., the clear difference between the description of an image (MSCOCO) and a question (QQP). To diminish this bias we trained the SVM on all datasets per method except once per dataset and method.

\subsection{Future Work}
This paper highlights the need for increased attention to machine-generated paraphrases, their alignment with human-authored texts, and their implications for various applications. While the data for human paraphrasing is both quantitative and qualitative abundant, we emphasize the importance of developing high-quality machine-generated datasets in the paraphrase detection field. Future work could focus on creating such datasets to enhance the study of machine-generated paraphrases.

In our study, we observed that certain methods obtain higher/lower correlations with increasing/decreasing diversity raising questions for future research on the architectures of detection methods and their varying performance on semantically balanced and diverse corpora. Identifying the architectural parameters that significantly impact paraphrase detection performance can contribute to the development of new or modified architectures that excel in this task.

Additionally, an interesting idea to explore is the incorporation of adversarial training into paraphrase detection models. By introducing adversarially generated paraphrases during the training process, models may become more robust and better equipped to handle challenging and unseen paraphrasing scenarios. This approach can lead to improved performance on diverse and complex datasets, enabling the development of more accurate and reliable paraphrase detection systems.

\clearpage


{
\small
\bibliography{anthology,custom}
\bibliographystyle{plainnat}
}

\clearpage

\appendix

\section{Reporting AI Usage For This Work}

We follow \citet{WahleRMM23} to report AI usage throughout the crafting of this scientific work.
Card \ref{card:this-paper} shows the report.

\input{card}

\end{document}

%% file: tables/method_comparison.tex
\begin{table*}[h]
    \centering
    \caption{\label{tab:table-detection-methods}
    Comparison of detection methods. The output of each method is used for training the SVM.
    }
    \renewcommand{\arraystretch}{1.5}
    \begin{tabular}{lp{7.5cm}l}
        \toprule
        \textbf{Detection Method} & \textbf{Description} & \textbf{Output}\\
        \midrule
        TF-IDF & Calculating the TF-IDF vector representation of both texts and taking their cosine similarity as result. & Similarity Score\\
        N-Gram & Generating the texts 3-Grams and computing their similarity score \citep{Kond}. & Similarity Score\\
        Fuzzy & Uses the Levenshtein distance in a correlation matrix to calculate the similarity between two texts \citep{WenYenFuzzy}. & Similarity Score\\
        GloVe & Represents each text within a global word vector space \citep{pennington-etal-2014-glove}. & Vectorized Texts\\
        Fasttext &  An updated variant of Word2Vec that represents each bag as character n-grams \citep{fasttext}. & Vectorized Texts\\
        BERT & A transformer model that considers the left and right context in all layers \citep{BERT}. & Embedded Texts\\
        T5 & A text-to-text transfer transformer built for translations and similar tasks \citep{T5}. & Embedded Texts\\
        \bottomrule
    \end{tabular}
\end{table*}

%% file: tables/parsed_datasets.tex
\begin{table}[htb]
  \caption{An executive summary of the parsed datasets.}
  \label{tab:parsed-datasets}
  \centering
  \begin{tabular}{llllllll}
    \toprule
    Dataset & Original Pairs & Filtered & Duplicates & Supplements & Our Data & Train & Test \\
    \midrule
    ETPC & 5,801  & 0 & 0 & 0 & 5,801 & 4k & 1,801 \\
    TURL & 56,787 & 2,260 & 1,061 & 0 & 10k & 4k & 6k \\
    QQP & 404,351 & 0 & 52 & 0 & 10k & 4k & 6k \\
    APH & 5,007 & 32 & 403 & 0 & 4,572 & 4k & 572 \\
    ParaSCI & 350,044 & 8 & 941 & 10k & 20k & 4k & 16k \\
    MSCOCO & 123,287 & 2 & 1 & 0 & 10k & 4k & 6k \\
    SaR & 2,038 & 0 & 0 & 0 & 2,038 & 1,486 & 552 \\
    \midrule
    SAv2 & 167,689 & 4,082 & 5 & 10k & 20k & 4k & 16k \\
    MPC & 163,715 & 35 & 0 & 10k & 20k & 4k & 16k \\
    ParaNMT & 51,409,584 & 176 & 10 & 10k & 20k & 4k & 16k \\
    APT & 138,004 & 0 & 3 & 0 & 10k & 4k & 6k \\
    PAWSWiki & 65,401 & 69 & 1,813 & 0 & 10k & 4k & 6k \\
    \bottomrule
  \end{tabular}
\end{table}

%% file: tables/gs_params.tex
\begin{table}[tbh]
\centering
\caption{Grid Search Space\label{tab:table-gs-params}}
\begin{tabular}[h]{ll}
    \toprule
    \textbf{Parameter} & \textbf{Range} \\
    \midrule
    kernel & linear, radial bases function, polynomial\\
    gamma &  0.01, 0.001, 0.0001, 0.0001\\
    polynomial degree &  1, 2, 3, 4, 5, 6, 7, 8, 9\\
    C & 1, 10, 100\\
    \bottomrule
\end{tabular}
\end{table}

%% file: tables/mean_cos_distances_embedded.tex
\begin{tabular}[b]{lll}
    \toprule
    \textbf{Dataset} & \textbf{Paraphr.} & \textbf{Orig.} \\
    \midrule
    ETPC & 0.93 & 0.90\\
    APH & 0.87 & 0.80\\
    TURL & 0.86 & 0.74\\
    QQP & 0.92 & 0.85\\
    MSCOCO & 0.84 & -\\
    ParaSCI & 0.93 & -\\
    SaR & 0.97 & -\\
    \midrule
    SAv2 & 0.93 & -\\
    MPC & 0.99 & -\\
    ParaNMT & 0.84 & -\\
    APT & 0.97 & 0.90\\
    PAWSWiki & 0.98 & 0.98\\
    \bottomrule
\end{tabular}
\label{tab:table-mean-cos-distance-embedded}

%% file: tables/methods_per_dataset_eval.tex
\begin{table*}
    \centering
    \caption{\label{tab:table-methods-per-dataset-eval}
    F1-score of detection methods per dataset. The \colorbox{green}{best} (green) and \colorbox{red}{worst} (red) results per dataset are highlighted.
    }
    \begin{tabular}{l|lllllll|l}
        \toprule
        \textbf{Dataset} & \textbf{TF-IDF} & \textbf{N-Gram} & \textbf{Fuzzy} & \textbf{GloVe} & \textbf{Fasttext} & \textbf{BERT} & \textbf{T5} & \textbf{Avg.}\\
        \midrule
        ETPC & \colorbox{green}{0.75} & 0.68 & 0.73 & \colorbox{red}{0.61} & 0.68 & 0.74 & 0.70 & 0.70\\
        APH & 0.69 & \colorbox{red}{0.44} & 0.54 & 0.71 & 0.70 & \colorbox{green}{0.85} & 0.81 & 0.68\\
        TURL & \colorbox{green}{0.70} & 0.45 & 0.52 & 0.33 & \colorbox{red}{0.18} & 0.63 & 0.55 & \colorbox{red}{0.48}\\
        QQP & \colorbox{green}{0.72} & 0.59 & 0.64 & 0.43 & \colorbox{red}{0.26} & 0.69 & 0.55 & 0.55\\
        MSCOCO & 0.23 & \colorbox{red}{0.09} & 0.13 & 0.87 & 0.86 & \colorbox{green}{0.95} & 0.93 & 0.58\\
        ParaSCI & 0.48 & \colorbox{red}{0.36} & 0.41 & 0.88 & 0.86 & \colorbox{green}{0.96} & 0.93 & 0.70\\
        SaR & 0.77 & \colorbox{red}{0.80} & 0.80 & 0.84 & 0.83 & \colorbox{green}{0.94} & 0.91 & \colorbox{green}{0.84}\\
        \midrule
        SAv2 & 0.76 & \colorbox{red}{0.60} & 0.64 & 0.78 & 0.61 & \colorbox{green}{0.94} & 0.86 & 0.74\\
        MPC & \colorbox{red}{0.78} & 0.84 & 0.82 & 0.87 & 0.84 & \colorbox{green}{0.95} & 0.93 & \colorbox{green}{0.86}\\
        ParaNMT & 0.53 & \colorbox{red}{0.47} & 0.52 & 0.73 & 0.78 & \colorbox{green}{0.95} & 0.93 & 0.70\\
        APT & 0.71 & \colorbox{green}{0.76} & 0.75 & \colorbox{red}{0.54} & 0.56 & 0.71 & 0.66 & 0.67\\
        PAWSWiki & 0.67 & 0.68 & \colorbox{green}{0.68} & 0.53 & \colorbox{red}{0.45} & 0.54 & 0.50 & \colorbox{red}{0.58}\\
        \bottomrule
    \end{tabular}
\end{table*}

%% file: tables/gini_coefficients.tex
\begin{tabular}[h]{ll}
    \toprule
    \textbf{Method} & \textbf{Gini} \\
    \midrule
    TF-IDF & \colorbox{red}{0.10}\\
    N-Gram & 0.13\\
    Fuzzy & 0.13\\
    GloVe & 0.22\\
    Fasttext & 0.17\\
    BERT & \colorbox{green}{0.36}\\
    T5 & 0.32\\
    \bottomrule
\end{tabular}

%% file: card.tex
{\sffamily
    \centering
    \tcbset{colback=white!10!white}
    \begin{tcolorbox}[
        title={\large \vspace{3mm} \textbf{\textit{Paraphrase Detection: Human vs. Machine Content}\vspace{3mm}}},
        breakable,
        boxrule=0.7pt,
        width=\textwidth,
        center,
        skin=bicolor,
        before lower={\footnotesize{AI Usage Card v1.0 \hfill \url{https://ai-cards.org} \hfill \href{https://ai-cards.blind-review.com/whitepaper.pdf}{PDF} | \href{https://ai-cards.blind-review.com/whitepaper.bib}{BibTeX}}},
        segmentation empty,
        halign lower=center,
        collower=white,
        colbacklower=tcbcolframe]
            
        \footnotesize{
            \begin{longtable}{p{.25\textwidth} p{.33\textwidth} p{.33\textwidth}}
              {\color{LightBlue} \MakeUppercase{Correspondence(s)}} \newline Jonas Becker
              & {\color{LightBlue} \MakeUppercase{Contact(s)}} \newline \href{mailto:becker@gipplab.org}{becker@gipplab.org}
              & {\color{LightBlue} \MakeUppercase{Affiliation(s)}} \newline University of Göttingen
              \\\\
              & {\color{LightBlue} \MakeUppercase{Project Name}} \newline Paraphrase Detection: Human vs. Machine Content
              & {\color{LightBlue} \MakeUppercase{Key Application(s)}} \newline Paraphrase Detection
              \\\\
              {\color{LightBlue} \MakeUppercase{Model(s)}} \newline ChatGPT
              & {\color{LightBlue} \MakeUppercase{Date(s) Used}} \newline 2023-03-14
              & {\color{LightBlue} \MakeUppercase{Version(s)}} \newline GPT-3.5, GPT-4\\\\
              \cmidrule{2-3}\\
      
              {\color{LightBlue} \MakeUppercase{Ideation}} \newline   
              & {\color{LightBlue} \MakeUppercase{Generating ideas, outlines, and workflows}} \newline Not used 
              & {\color{LightBlue} \MakeUppercase{Improving existing ideas}} \newline Not used. \\\\
              & {\color{LightBlue} \MakeUppercase{Finding gaps or compare aspects of ideas}} \newline Not used \\\\
              
              {\color{LightBlue} \MakeUppercase{Literature Review}} \newline    
              & {\color{LightBlue} \MakeUppercase{Finding literature}} \newline Not used
              & {\color{LightBlue} \MakeUppercase{Finding examples from known literature}} \newline Not used \\\\
              & {\color{LightBlue} \MakeUppercase{Adding additional literature for existing statements and facts}} \newline Not used
              & {\color{LightBlue} \MakeUppercase{Comparing literature}} \newline Not used \\\\
              \cmidrule{2-3}\\
      
              {\color{LightBlue} \MakeUppercase{Methodology}} \newline    
              & {\color{LightBlue} \MakeUppercase{Proposing new solutions to problems}} \newline Not used
              & {\color{LightBlue} \MakeUppercase{Finding iterative optimizations}} \newline Not used \\\\
              & {\color{LightBlue} \MakeUppercase{Comparing related solutions}} \newline Not used. \\\\
              
              {\color{LightBlue} \MakeUppercase{Experiments}} \newline    
              & {\color{LightBlue} \MakeUppercase{Designing new experiments}} \newline Not used
              & {\color{LightBlue} \MakeUppercase{Editing existing experiments}} \newline Not used \\\\
              & {\color{LightBlue} \MakeUppercase{Finding, comparing, and aggregating results}} \newline Not used \\\\
              \cmidrule{2-3}\\
      
              {\color{LightBlue} \MakeUppercase{Writing}} \newline ChatGPT   
              & {\color{LightBlue} \MakeUppercase{Generating new text based on instructions}} \newline Generated a first version of the abstract.
              & {\color{LightBlue} \MakeUppercase{Assisting in improving own content}} \newline Improving parts of the abstract, introduction, and discussion. \\\\
              & {\color{LightBlue} \MakeUppercase{Paraphrasing related work}} \newline Not used 
              & {\color{LightBlue} \MakeUppercase{Putting other works in perspective}} \newline Not used \\\\
              
              {\color{LightBlue} \MakeUppercase{Presentation}} \newline    
              & {\color{LightBlue} \MakeUppercase{Generating new artifacts}} \newline Not used
              & {\color{LightBlue} \MakeUppercase{Improving the aesthetics of artifacts}} \newline Not used \\\\
              & {\color{LightBlue} \MakeUppercase{Finding relations between own or related artifacts}} \newline Not used \\\\
              \cmidrule{2-3}\\
              {\color{LightBlue} \MakeUppercase{Coding}} \newline    
              & {\color{LightBlue} \MakeUppercase{Generating new code based on descriptions or existing code}} \newline Not used
              & {\color{LightBlue} \MakeUppercase{Refactoring and optimizing existing code}} \newline Not used \\\\
              & {\color{LightBlue} \MakeUppercase{Comparing aspects of existing code}} \newline Not used \\\\
              
              {\color{LightBlue} \MakeUppercase{Data}} \newline    
              & {\color{LightBlue} \MakeUppercase{Suggesting new sources for data collection}} \newline Not used 
              & {\color{LightBlue} \MakeUppercase{Cleaning, normalizing, or standardizing data}} \newline Not used  \\\\
              & {\color{LightBlue} \MakeUppercase{Finding relations between data and collection methods}} \newline Not used  \\\\
              \cmidrule{2-3}\\
      
              {\color{LightBlue} \MakeUppercase{Ethics}} \newline ChatGPT   
              & {\color{LightBlue} \MakeUppercase{What are the implications of using AI for this project?}} \newline Enabling the comparison of paraphrase detection for newly generated content by language models.
              & {\color{LightBlue} \MakeUppercase{What steps are we taking to mitigate errors of AI for this project?}} \newline Careful evaluation of any generated content from ChatGPT. \\\\
              & {\color{LightBlue} \MakeUppercase{What steps are we taking to minimize the chance of harm or inappropriate use of AI for this project?}} \newline Documentation of the steps and suggested text in the scientific document.
              & {\color{LightBlue} \MakeUppercase{The corresponding authors verify and agree with the modifications or generations of their  used AI-generated content}} \newline Yes \\
            \end{longtable}
        }
        \tcblower
    \end{tcolorbox}
    \setcounter{table}{0}
    \captionof{table}{AI Usage Card for this project.}
    \label{card:this-paper}
}